Ari Goodman, James Hing, Gurpreet Singh, Ryan O'Shea
Naval Air Warfare Center Aircraft Division Lakehurst

# Computer Vision for Carriers: PATRIOT

## Abstract

Deck tracking performed on carriers currently involves a team of sailors manually identifying aircraft and updating a digital user interface called the Ouija Board. Improvements to the deck tracking process would result in increased Sortie Generation Rates, and therefore applying automation is seen as a critical method to improve deck tracking. However, the requirements on a carrier ship do not allow for the installation of hardware-based location sensing technologies like Global Positioning System (GPS) sensors. PATRIOT (Panoramic Asset Tracking of Real-Time Information for the Ouija Tabletop) is a research effort and proposed solution to performing deck tracking with passive sensing and without the need for GPS sensors. PATRIOT is a prototype system which takes existing camera feeds, calculates aircraft poses, and updates a virtual Ouija board interface with the current status of the assets. PATRIOT would allow for faster, more accurate, and less laborious asset tracking for aircraft, people, and support equipment. PATRIOT is anticipated to benefit the warfighter by reducing cognitive workload, reducing manning requirements, collecting data to improve logistics, and enabling an automation gateway for future efforts to improve efficiency and safety. The authors have developed and tested algorithms to perform pose estimations of assets in real-time including OpenPifPaf, High-Resolution Network (HRNet), HigherHRNet (HHRNet), Faster R-CNN, and in-house developed encoder-decoder network. The software was tested with synthetic and real-world data and was able to accurately extract the pose of assets. Fusion, tracking, and real-world generality are planned to be improved to ensure a successful transition to the fleet.

## Introduction
Tracking of aircraft, people, and support equipment to update the Ouija board is performed manually through visual inspection by several sailors. This method leads to data delays, errors, and reduced situational awareness. This task is labor intensive, requires intense focus for extended periods of time, and requires multiple personnel stationed during all flight operations. The Program of Record has identified a need for a technology to enable automatic tracking of assets on deck for over a decade. PATRIOT is a new solution allow for faster, more accurate, and less laborious asset tracking, as well as to enable future efforts focused on optimization, logging, and robust tracking.

In this work, the detection, classification, and pose estimation components for PATRIOT are presented along with experiments to quantify their training and performance.

## Background
PATRIOT's main tasking is broken down into three tasks: detection, classification, and pose estimation. The detection task involves identifying regions in the image and associating them with objects of interest. The classification task involves applying a class label to each object detected in the image. The pose estimation task involves estimating the 3D position and orientation of an object from its 2D image. There are numerous challenges in these tasks including perspective distortion, variable lighting conditions, noisy data, and occlusions [1].

Detection and classification algorithms have been the focus of many computer vision efforts. There are several freely available state-of-the-art open-source solutions such a YOLO [2] and Faster R-CNN [3]. Although well-trained, freely available models may need retraining, also called transfer learning or fine-tuning, to meet performance requirements on new datasets [4,5].

Although pose estimation is not as well studied as detection and classification, several methods have been developed for estimating the pose of people and aircraft. Many state-of-the-art methods in single object pose estimation break down the process into two main steps. First, the algorithms identify the location of key features of objects, such as the nose or wingtips of an aircraft. This step is referred to as the feature detection or keypoint detection step. Then, the pose is calculated using Perspective-n-Point (PnP) solvers [6]. PnP algorithms traditionally use the known 3D points, corresponding image points, and camera parameters to estimate the real-world pose of the known object. PnP algorithms attempt to minimize the error between the projected points in 2D and the measured points in the image [7].

Three state-of-the-art methods for keypoint detection are HRNet [8], HHRNet [9], OpenPifPaf [10].

There are pros and cons for each approach. In general, top-down methods like HRNet tend to be more accurate when dealing with large changes in scale of objects in an image because the bounding box step essentially normalizes the scale. Bottom-up methods, on the other hand, are more accurate when dealing with overlapping objects. In terms of processing speed, because bottom-up methods like HHRNet and OpenPifPaf don't have the additional step of running an object detector first, their prediction times can be faster [9].

All three algorithms have been demonstrated to achieve strong performance on a variety of benchmarks and have been used in a wide range of applications. However, their performance can vary based on the task and dataset, so it was unclear how they would perform with PATRIOT's datasets [8,9,10].

An alternative approach to pose estimation is to use direct linear transforms (DLT) and decoders. DLT works by projecting points onto an image place using known intrinsic and extrinsic camera parameters. DLT can also be used to estimate the camera parameters using two sets of known 3D and 2D points. Encoder-decoder networks are a type of deep learning architecture, but are traditionally smaller than the aforementioned HRNet, HHRNet, and OpenPifPaf networks, and therefore may require less data to train. Encoder-decoder networks consist of two main parts: an encoder, which processes the input data and encodes it into a compact representation, and a decoder, which takes the encoded representation and converts it back into the desired output. In the context of PATRIOT, an encoder-decoder network could be used to estimate the orientation of aircraft in images. The encoder would process the input image and extract relevant features, such as the shape and texture of the aircraft, while the decoder would learn to use these features to estimate the orientation of the aircraft.

## Methodology

Four pipelines were developed, each which are distinguished by their respective use of HRNet, Higher HRNet, OpenPifPaf, and a decoder. Each pipeline was designed to take in an image and output a list of objects with pose, class, and confidences.

In this work, an experiment is included comparing all pipelines on a common synthetic dataset. Another experiment compares three OpenPifPaf models on real-world data; one model was trained on synthetic data, another on real-world, and a final model was trained on both real-world and synthetic data.

### Frameworks
Four frameworks were used to train and evaluate candidate algorithms for pose estimation.

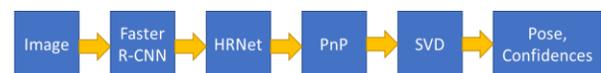
Figure 1: HRNet Framework

In the first approach shown in Figure 1, an image was passed to a detection and classification component, Faster R-CNN. The output from Faster R-CNN and the image was passed to a keypoint detection model, HRNet. The keypoint detection model produced a list of keypoint locations found for the object in the image. Next, the keypoints and class were passed into a PnP solver to estimate the real-world position of all the keypoints [11]. Finally, the keypoints and their labels were passed to a Singular Value Decomposition (SVD) solving algorithm to minimize the error between translating, rotating, and scaling a known point cloud set of keypoints to the estimated keypoints.

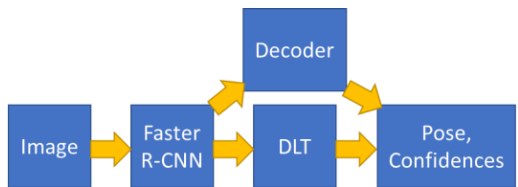
Figure 2: Decoder Framework

In the second approach shown in Figure 2, the output from the detector and classifier, Faster R-CNN, was instead passed to a decoder network and a Direct Linear Transform (DLT) component. The decoder network was trained to estimate the orientation of a single object in the image. It was assumed that the object was on the carrier deck (Z=0). The DLT component estimated the real-world X and Y position of the object under the assumption the object's pixel position was at the center of Faster R-CNN's bounding box. Combining the DLT's position and the decoder's orientation estimate formed the final pose estimate.

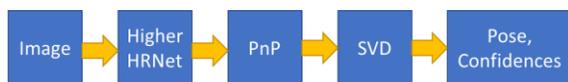
Figure 3: Higher HRnet Framework

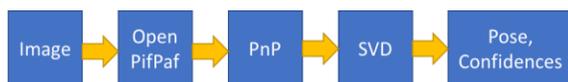
Figure 4: OpenPifPaf Framework

In the third and fourth approaches shown in Figures 3 and 4, an image was passed into bottom-up algorithms HHRNet or OpenPifPaf. These algorithms directly estimated the sets of keypoints for each object. Then, the keypoints and class were passed into the same PnP and SVD components as in the first framework.

**Datasets**

A combination of real-world data and synthetic data was used to develop and test the pose estimation pipeline.

A small assortment of video imagery previously recorded during the EATS TEMPALT aboard the U.S.S. Truman, CVN-75 was labeled and used as a real-world dataset. A panoramic camera assembly and two additional fixed cameras mounted on the island provided full video coverage of the flight deck. A real-world dataset was created with 4,964 annotated images with an additional 553 used for validation and testing.

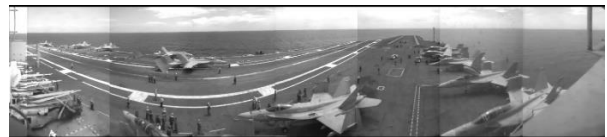
Figure 5: Example of Panoramic Camera Footage

Synthetic data was used in this project because only a small amount of real-world data was available. In addition, synthetic data allowed for the quick and accurate labeling of images, as well as full control over images that would be difficult to capture, such as unique aircraft configurations, lighting, or weather.

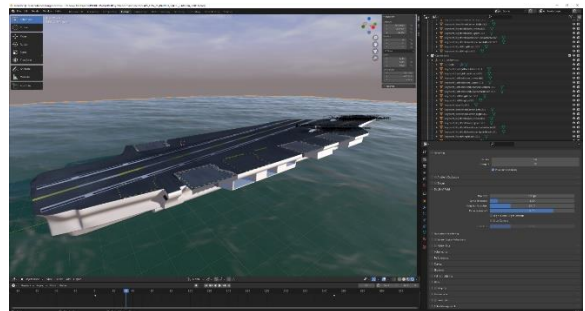
Figure 6: Example of Blender Environment

Blender 3.3 was used as a tool to render synthetic scenes for PATRIOT. Blender is a free and open-source 3D creation suite [12]. Freely available open-source models of the aircraft carrier and F-18 were used. The carrier model was created by Alexdark [13] and the F/A-18 Hornet was created by KuhnIndustries [14]. An example of the synthetic environment can be seen in Figure 6 and a result of a synthetic rendered camera image is shown in Figure 7. The rendered camera image is a simulated version of the panoramic camera that is installed on the actual carrier. It is made up of a series of 5 cameras co-located next to each other on the carrier island.

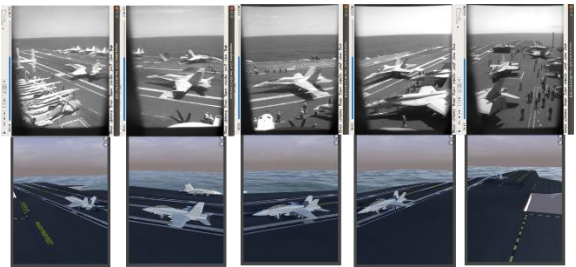

Figure 7: Blender Rendered Camera vs Real-World Footage

10,000 synthetic images were generated: 8,000 for training and 2,000 for testing and validation.

17 key features on the aircraft were identified to train the keypoint detection methods based on geometric location. Future work could address the identification of features that are most easily detected. The 17 keypoints are highlighted in a skeleton in Figure 8.

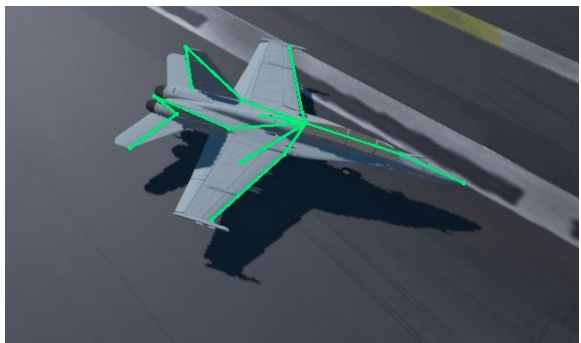

Figure 8: Labeled Aircraft Skeleton

### Faster R-CNN

Faster R-CNN is an object detector model that uses a convolutional neural network (CNN) based architecture. The Faster R-CNN architecture detected and classified aircraft in images. It output a list of bounding boxes, classes, and confidences. Figure 9 shows Faster R-CNN working on real-world data.

For Faster R-CNN, the training parameters were: train_batch_size = 1, num_epochs = 10, lr = 0.005, momentum = 0.9, weight_decay = 0.005.

### Direct Linear Transform

The DLT algorithm projects points onto an image place using known intrinsic and extrinsic camera parameters. Given a 2D point, DLT estimates the corresponding camera ray in 3D. In PATRIOT's dataset, it was assumed that objects are on the carrier deck, and therefore have a known Z height. Therefore, DLT was used to solve for the real-world X, Y, and Z position of a 2D point in the image. An example image with Faster R-CNN working with the DLT is shown in Figure 9.

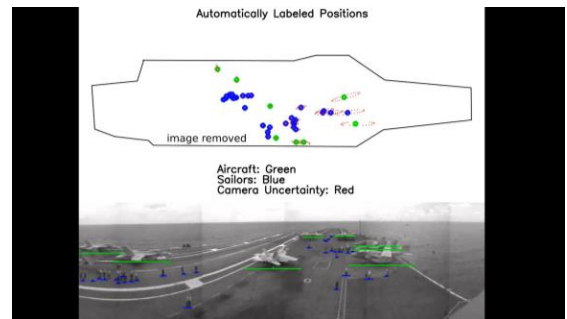

Figure 9: Faster R-CNN and DLT on Real-World Data

DLT was also used to estimate the camera parameters using two sets of corresponding, known 3D and 2D points. DLT attempted to find the camera parameters that minimized the error to project the points from one space to the other. The authors mapped dozens of known points from the ship environment to the 2D image to calibrate the cameras.

### Encoder-Decoder Network

The encoder-decoder network utilized a standard convolution autoencoder structure with the addition of a second yaw estimation head. The encoder portion of the network learned to use convolution layers to extract information from the image and distill it down into an encoded representation vector. The decoder portion of the network learned to use inverse convolution layers to reconstruct the input image from the encoded representation vector; the reconstruction loss between images was used as a measure of confidence. Part of the encoded representation vector is also used by the yaw estimation head to calculate the yaw of the object. The yaw estimation head was structured as a fully connected neural network that terminated in overlapping yaw range bins. Each yaw bin represented a set range of rotations that an object could possibly take on. After the most confident bin is selected, the network regressed the final yaw of the object based on the pre-defined bin centers. An overview of the algorithm is shown in Figure 10.

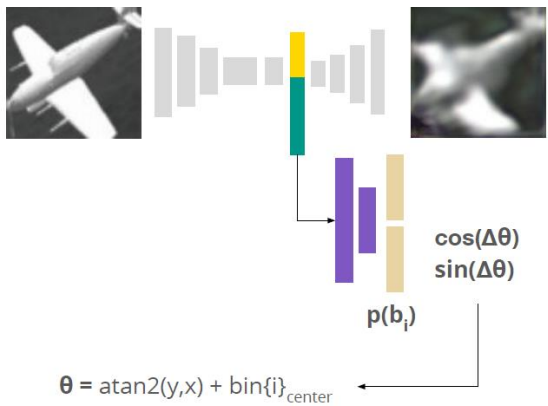

Figure 10: Encoder-Decoder Network Diagram

Three loss functions were used during training and could be weighted to increase or decrease focus on specific tasks. The three loss functions were decoder image reconstruction loss, bin selection loss, and rotational offset loss. The ADAM optimizer was used in conjunction with the three loss functions to train the network for a set number of epochs [15].

The encoder-decoder's performance on a synthetic dataset is shown in Figures 11 and 12.

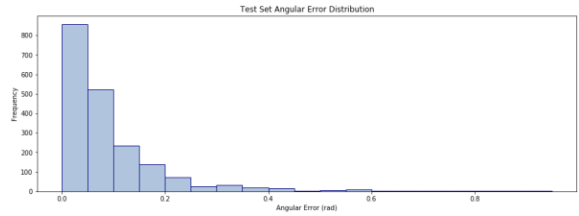

Figure 11: Encoder-Decoder Angular Error

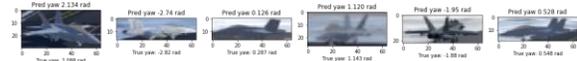

Figure 12: Selected Examples for Encoder-Decoder Orientation Estimation

### HRNet

High-Resolution Network (HRNet) is a top-down convolutional neural network (CNN) based architecture for computer vision tasks. The HRNet architecture was used to detect keypoints from the images provided by Faster R-CNN. It output a list of keypoint heatmaps. For each heatmap, the pixel with the highest heat value was selected as the keypoint location.

The training parameters used were: batch_size_per_gpu: 8, shuffle: true, begin_epoch: 0, end_epoch: 120, optimizer: adam, lr: 0.0005, lr_factor: 0.1, lr_step: - 90 – 110 wd: 0.0001, gamma1: 0.99, gamma2: 0.0, momentum: 0.9

### OpenPifPaf

OpenPifPaf is a bottom-up keypoint detection approach based on ResNet with two head networks. The OpenPifPaf architecture was used to detect keypoints from the entire image. An example of OpenPifPaf working on preliminary real-world data is shown in Figure 13.

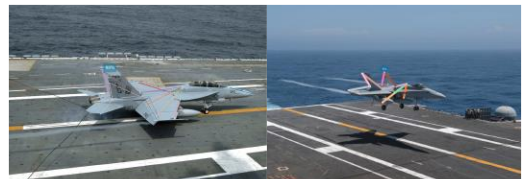

Figure 13: Selected OpenPifPaf Results on Real-World Data

The training parameters used were: --lr=0.0002 –momentum=0.95 –b-scale=5.0 –epochs=1000 –lr-warm-up-epochs=100 –batch-size=9 –loader-

workers=8 –val-interval=100 –weight-decay=1e-5

### Higher HRNet

HHRNet is also a bottom-up keypoint detection approach. The HHRNet architecture was used to detect keypoints from the entire image. A selected result quantifying its performance on real-world data is shown in Figure 14.

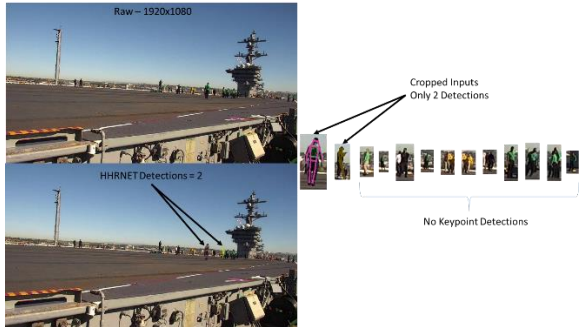

Figure 14: Selected Higher HRNet Results on Real-World Data

The training parameters used were:
loss: num_stages: 2, ae_loss_type: exp, with_ae_loss: [true, false], push_loss_factor: [0.001, 0.001], pull_loss_factor: [0.001, 0.001], with_heatmaps_loss: [true, true], heatmaps_loss_factor: [1.0, 1.0],
begin_epoch: 0, checkpoint: '', end_epoch: 300, gamma1: 0.99, gamma2: 0.0, images_per_gpu: 12, lr: 0.001, lr_factor: 0.1, lr_step: [200, 260], momentum: 0.9, nesterov: false, optimizer: adam, resume: false, shuffle: true, wd: 0.0001

### Evaluation Experiment

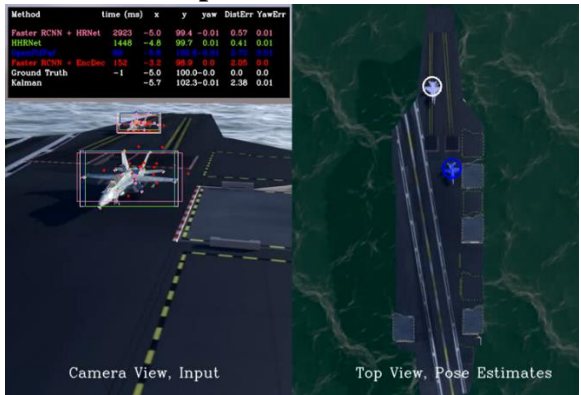

Figure 15: Image from Evaluation Experiment

An experiment was conducted to compare the various pose estimation methods. A synthetic environment in Blender was chosen for its controllability and ease of creation. A video was created in which two aircraft moved throughout the scene to test each algorithm when the aircraft were partially and fully occluded. An example image taken from the processed demonstration video is shown in Figure 15.

### How Useful is Synthetic Data?

Three OpenPifPaf models were evaluated on a real-world dataset. First, an OpenPifPaf model was trained on a synthetic Blender data set. Second, a model was trained on a subset of the real-world dataset. Third, the synthetically trained model was retrained on a subset of real-world data.

### Results

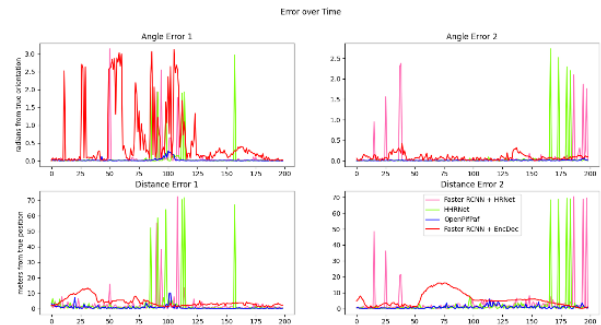

Figure 16: Graph of Error for Each Framework for Each Aircraft

| Name | % in Spec Plane 1 | % in Spec Plane 2 | Median time | Median Distance Error 1 | Median Angle Error 1 | Median Distance Error 2 | Median Angle Error 2 |
|---|---|---|---|---|---|---|---|
| Open PifPaf | 94% | 95% | 69ms | 0.4m | 0.3° | 0.4m | 0.3° |
| HHRNet | 87% | 94% | 729ms | 0.4m | 0.3° | 0.6m | 0.5° |
| Faster RCNN + HRNet | 90% | 83% | 170ms | 0.9m | 0.7° | 1.1m | 0.7° |
| Faster RCNN + EncDec | 12% | 18% | 2279ms | 3.2m | 11° | 4.8m | 3.7° |

Figure 17: Table of Error for Each Framework for Each Aircraft

The resulting accuracies can be seen in Figures 16 and 17. OpenPifPaf was the most accurate and fastest overall. All algorithms besides the decoder pipeline were in specification (within 1 m and 0.5 degrees) a majority of the time.

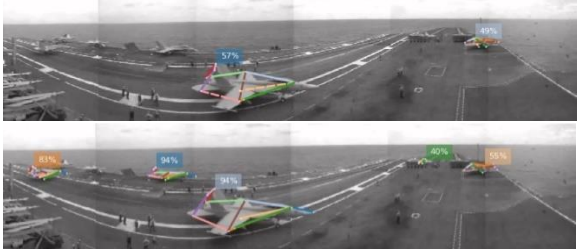

Figure 18: Top: Real-World Trained Model vs. Bottom: Synthetic + Real World Trained Model

Running the synthetic data trained OpenPifPaf model on real-world data did not generate consistent or useable results, however it did show some promise even though it had only been trained on synthetic data. OpenPifPaf only trained on real-world data had consistent and useable results, but the best results were from the model trained with synthetic and real-world data. As shown in Figure 18, there were more keypoint detections and the confidence level is higher. This experiment demonstrated that the combination of synthetic data with real-world data was helpful to improve keypoint detection.

## Conclusion

A pipeline and framework for detection, classification, and pose estimation of assets in real-world and synthetic data was successfully demonstrated. A keypoint detection model trained on synthetic data plus real-world data can detect keypoints in real-world footage.

A synthetic environment was demonstrated to be useful in performing experiments and supplementing real-world data when training pose estimation software.

The full pipeline and SOTA algorithms within meet requirements to detect, classify, and estimate the pose of aircraft in real-time when properly trained on the appropriate data. The best performing algorithm for speed and accuracy was OpenPifPaf.

The limited data set that was annotated for proof of concept in this project will be expanded to improve the performance of the machine learning models, as well as to allow for fusion and tracking

## Acknowledgements

The authors would like to acknowledge Vitaly Ablavsky from the University of Washington for his assistance in designing the encoder-decoder network. The authors would also like to acknowledge Ric Rey Vergara, Mark Blair, and Daniel Vidal for interfacing with Navy personnel.

Ari Goodman is the S&T AI Lead and a Robotics Engineer in the Robotics and Intelligent Systems Engineering (RISE) lab at Naval Air Warfare Center Aircraft Division (NAWCAD) Lakehurst. In this role he leads efforts in Machine Learning, Computer Vision, and Verification & Validation of Autonomous Systems. He received his MS in Robotics Engineering from Worcester Polytechnic Institute in 2017.

Dr. James Hing serves as the Branch Head of the Strategic Technologies Branch, NAWCAD Lakehurst, where he leads a team of 25 engineers, including 4 R&D Laboratories (IDATS, RISE, INSPIRE, ADHaTEC), in the evaluation, development, maturation, and transition of new and emerging technologies with application to Aircraft Launch Recovery and Support Equipment (ALRE/SE). He has 20 years of multidisciplinary expertise in the fields of computer vision, robotics, and autonomous systems. He received his PhD in Mechanical Engineering from Drexel University.

Gurpreet Singh is a Computer Scientist in the Robotics and Intelligent Systems Engineering (RISE) lab at Naval Air Warfare Center Aircraft Division (NAWCAD) Lakehurst. His areas of interest and expertise are Computer Vision and Deep Learning. He received his Master's of Science in Computer Science from Stevens Institute of Technology and Graduate Certificate in Robotics Engineering from University of Maryland.

Ryan O'Shea is a Computer Engineer in the Robotics and Intelligent Systems Engineering (RISE) lab at Naval Air Warfare Center Aircraft Division (NAWCAD) Lakehurst. His current work is focused on applying computer vision, machine learning, and robotics to various areas of the fleet in order to augment sailor capabilities and increase overall operational efficiency. He received a Bachelor's Degree in Computer Engineering from Stevens Institute of Technology.